\documentclass[a4paper,twoside]{article}

\usepackage{epsfig}
\usepackage{subcaption}
\usepackage{calc}
\usepackage{amssymb}
\usepackage{amstext}
\usepackage{amsmath}
\usepackage{amsthm}
\usepackage{multicol}
\usepackage{pslatex}
\usepackage{apalike}

\usepackage{bm}
\usepackage{multirow}
\usepackage{siunitx} 
\usepackage{graphicx}

\usepackage[ruled,vlined]{algorithm2e}

\usepackage{SCITEPRESS}     

\begin{document}

\title{\uppercase{Active Reinforcement Learning}
\subtitle{A Roadmap Towards Curious Classifier Systems for Self-Adaptation} }

\author{\authorname{Simon Reichhuber\sup{1}\orcidAuthor{0000-0001-8951-8962} and Sven Tomforde\sup{1}\orcidAuthor{0000-0002-5825-8915}}
\affiliation{\sup{1}Intelligent Systems, University of Kiel, Hermann-Rodewald-Str. 3, Kiel, Germany}
\email{\{sir,st\}@informatik.uni-kiel.de}
}


\abstract{Intelligent systems have the ability to improve their behaviour over time taking observations, experiences or explicit feedback into account. Traditional approaches separate the learning problem and make isolated use of techniques from different field of machine learning such as reinforcement learning, active learning, anomaly detection or transfer learning, for instance. In this context, the fundamental reinforcement learning approaches come with several drawbacks that hinder their application to real-world systems: trial-and-error, purely reactive behaviour or isolated problem handling. The idea of this article is to present a concept for alleviating these drawbacks by setting up a research agenda towards what we call ``active reinforcement learning'' in intelligent systems.}

\onecolumn \maketitle \normalsize \setcounter{footnote}{0} \vfill

\section{\uppercase{Introduction}}
\label{sec:introduction}

\noindent Information and communication technology faces a trend towards increasingly complex solutions, e.g., characterised by the laws of Moore \cite{Moo65} and Glass \cite{Gla02}. As a consequence, traditional concepts for design, development, and maintenance have reached their limits. Within the last decade, a paradigm shift in engineering such systems has been postulated that claims to master complexity issues by means of self-adaptation and self-organisation. Concepts and techniques emerged that move traditional design-time decisions to runtime and from the system engineer to the systems themselves. As a result, intelligent and autonomously acting systems are targeted, with the self-adapting and self-organising (SASO) systems domain serving as an umbrella for several research initiatives focusing on these issues, including Organic Computing \cite{MST17}, Autonomic Computing \cite{KC03}, Interwoven Systems \cite{THS14}, or Self-aware Computing Systems \cite{Kounev2017}.

The basic idea is in all cases that individual systems react autonomously to changing conditions, find appropriate reactions, and optimise this process over time---resulting in intelligent system behaviour. For the remainder of this article, we define such an ``intelligent system'' (according to \cite{TomfordeSM17}) as a computing system that achieves or maintains a certain level of performance, even when operating in environments that change over time and even if it is exposed to disturbances or emergent situations. Such an intelligent system is autonomously alerting its own behaviour with the goal to improve it over time.

A keystone in this definition is the ability of an intelligent system to learn autonomously at runtime \cite{DAngeloGGGNPT19}. This means that approaches based on massive training data or continuous feedback/supervision by users are not feasible. In turn, the system has to figure out what to do in which situation: the classic reinforcement learning (RL) paradigm combined with further mechanisms from the domain of machine learning such as anomaly detection, transfer learning, or collaborative learning \cite{DAngeloGGGNTP20}.

Several approaches have been presented where varying RL techniques are used for enabling self-improving runtime adaptation and organisation. However, these approaches come with several drawbacks that hinder their application to real-world systems: trial-and-error, purely reactive behaviour, isolated problem handling, etc. The idea of this article is to present a concept for alleviating these drawbacks by setting up a research agenda towards what we call ``active reinforcement learning'' in intelligent systems.

The remainder of this article is organised as follows: Section~\ref{sec:back} describes the technical background of this article. This includes a basic understanding of how intelligent systems are designed, fundamentals for reinforcement learning and its utilisation in intelligent systems, as well as a brief introduction of the key ideas of active learning. Afterwards, Section~\ref{sec:approach} describes our concept for active reinforcement learning, which results in the definition of a research road-map presented in Section~\ref{sec:roadmap}. Finally, Section~\ref{sec:conclusion} summarises the article and gives an outlook to future work.

\section{\uppercase{Background}}
\label{sec:back}

\noindent This section briefly describes basic concepts necessary for understanding the remainder of this article. We initially introduce our model of an intelligent system with an emphasis on learning autonomously at runtime. Afterwards, we explain the basic RL paradigm and the most popular variant as used in intelligent systems. Finally, we summarise the concept of active learning (AL) with a special focus on stream data consideration as most prominent variant for utilisation in intelligent systems.

\subsection{System model}
\label{sec:sysModel}

\begin{figure}[htb]
	\centering
	\includegraphics[width=0.99\linewidth]{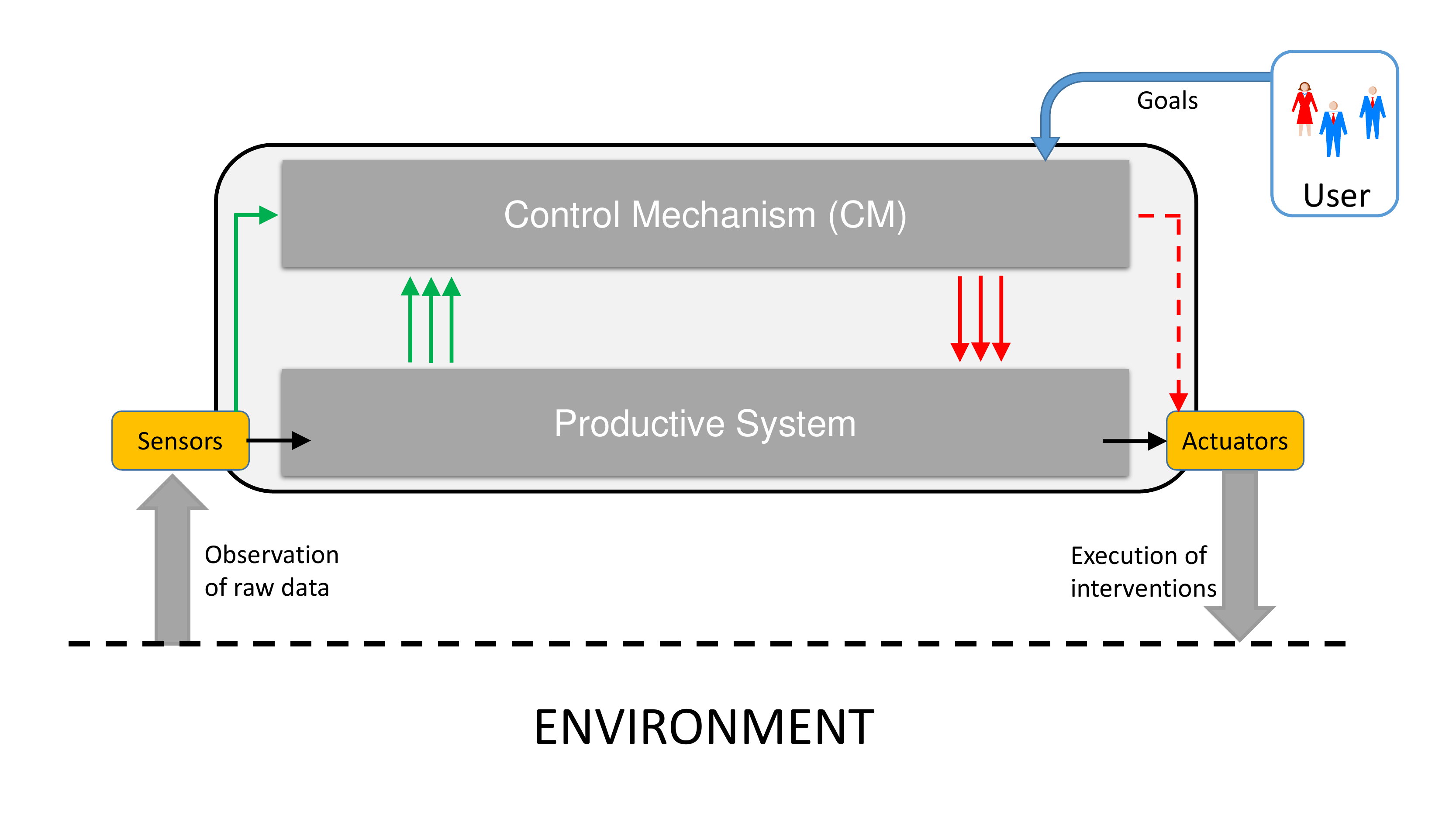}
	\captionof{figure}{Schematic illustration of a subsystem $a_i$.}\label{fig:SysPart}
\end{figure}

We assume a SASO system $S$ to consist of a potentially large set $A$ of autonomous subsystems $a_i$ in a virtual, physical or hybrid environment. We refer to the term \textit{(sub)system} using the terminology from the Organic Computing domain~\cite{MST17}, and, for a better readability, omit the \textit{sub} if it is clear from the context that not the overall system is meant (synonyms are \textit{entity} or \textit{agent}). Each $a_i$ is equipped with sensors and actuators. Internally, each $a_i$ distinguishes between a productive system part (\textit{PS}, responsible for the basic purpose of the system) and a control mechanism (CM, responsible for controlling the behaviour of the PS and deciding about relations to other subsystems). This corresponds to the separation of concerns between \textit{System under Observation and Control} (SuOC) and \textit{Observer/Controller} tandem in the terminology of Organic Computing (OC) \cite{TP+11} or \textit{Managed Resource} and \textit{Autonomic Manager} in terms of Autonomic Computing \cite{KC03}. Figure~\ref{fig:SysPart} illustrates this concept with its input and output relations. 

Each subsystem in the overall system can assume different configurations. Such a configuration typically consists of different components. We define the entire configuration space of a subsystem $a_i$ as Cartesian product $C_i=c_{i1}\times\dots\times c_{im}$, where $c_{ij}$ are the components of the configuration. The user guides the behaviour of $a_i$ using a utility function $U$ and (in most cases) does not intervene at the decision level: Actual decisions are taken by the productive system and the CM. We model each subsystem to act \textit{autonomously}, i.e., there are no control hierarchies in the overall system. Consider, e.g., a router $a_1$ in a computer network as an illustrating example following the ideas of \cite{THH10-a}. It can take varying configurations into account, such as the processed network protocol or parameter settings \cite{TS+09}. E.g., an interval $c_{11} = [0,100]$ for the timeout parameter in seconds and the set $c_{12}=\{1,2,\dots,16\}$ for the buffer size in kilobyte. The entire configuration space of system $a_1$ would then be $C_1=[0,100]\times\{1,2,\dots,16\}$.

Besides the configuration space, we consider a local reward. In particular, each subsystem estimates the success of its decisions at runtime---as a response to actions taken before. This is realised based on a feedback mechanism, with feedback possibly stemming from the environment of the subsystem (i.e., direct feedback) or from manual reward assignments (i.e., indirect feedback). This resembles the classic reinforcement model~\cite{SuttonB1998}, where the existence of such a reward is one of the basic assumptions.

\subsection{Reinforcement learning for self-adaptation in intelligent systems}
\label{sec:bg:rl}
\begin{figure}[ht!]
\centering
\includegraphics[width=0.99\columnwidth]{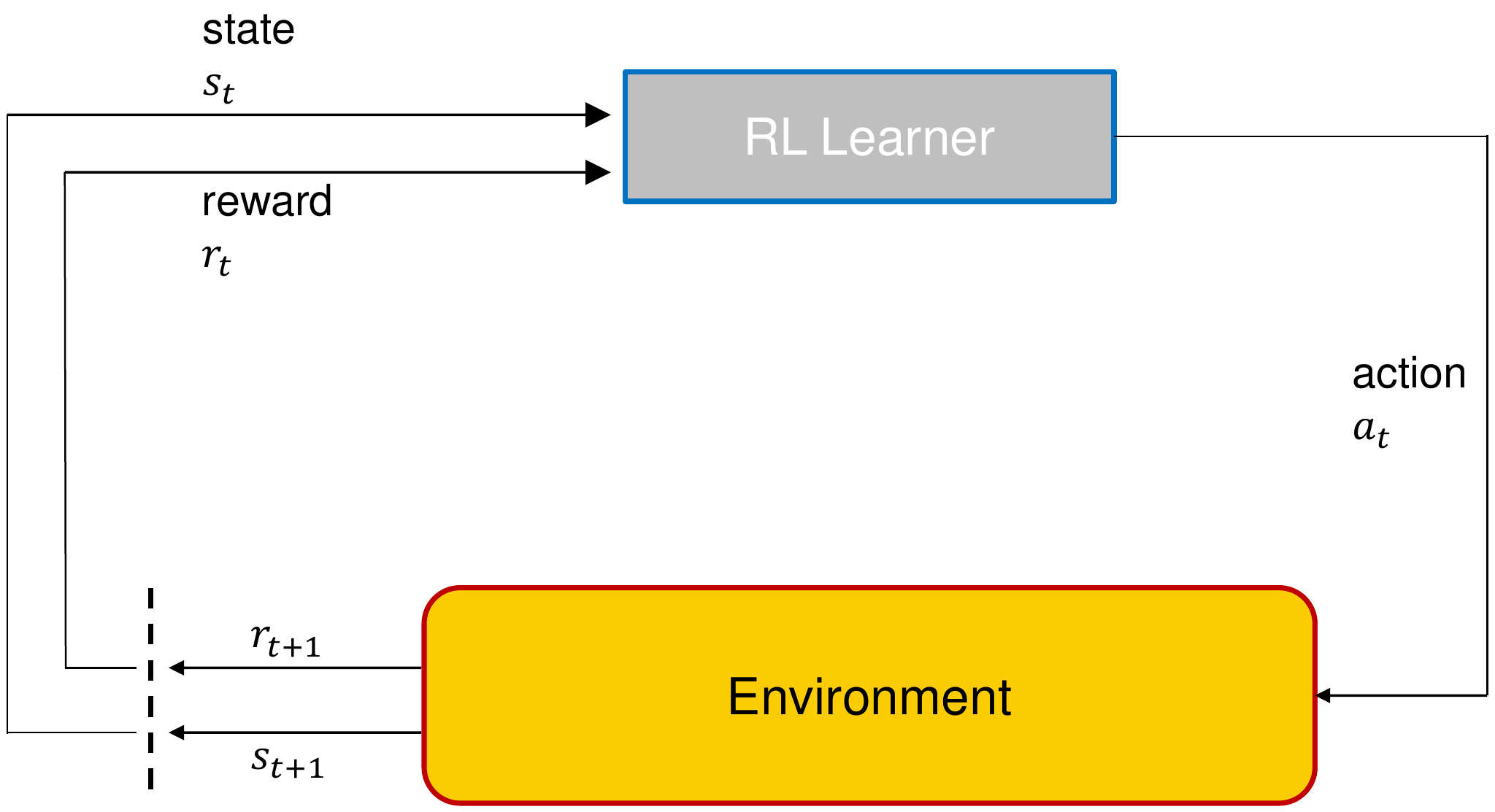}
\caption{The basic RL model}
\label{fig:RLmodel}
\end{figure}

The basic reinforcement learning (RL) paradigm relies on a continuous interplay of learner and environment. The control mechanism of an intelligent system continuously monitors the environmental conditions and thereby perceives the current state of the environment ($s_t$). The index $t$ refers to the point in time of the perception indicating that the RL loop is performed in discrete time steps. Based on this state description, the RL learner decides about the next action ($a_t$) that manipulates the environment (e.g., a change of parameters of a controlled productive system). This results in a possibly changed state of the environment. Hence, in the next time step $(t+1)$ an updated state ($s_{t+1}$) is presented to the learner. In addition, the learner receives a feedback signal ($r_{t+1}$) that quantifies the success of the action (if an immediate reward is possible) or is used to assess the 'value' of the current situation. Figure~\ref{fig:RLmodel} illustrates this process.

Several different RL techniques are available in literature. However, the utilisation of RL in intelligent system (especially for adaptive self-configuration, for instance) come with a set of requirements rendering most of the techniques not applicable--the most important are:

\begin{enumerate}
    \item Decisions are taken based on sensor information, which is typically available in terms of time-series of different real-valued attributes. As a consequence, the RL technique has to deal with real values.
    \item An intelligent system typically uses more than just a few sensor values, i.e. it has to master a situation space $\mathbb{R}^d$ with $d$ defining the number of dimensions of this space.
    \item The efficiency of the RL technique depends directly on the number of possible choices, since learning requires testing different possibilities. Consequently, a simple list of situation-action mappings is not feasible and the learner needs the ability to generalise.
\end{enumerate}

\begin{figure*}[ht!]
\centering
\includegraphics[width=0.85\textwidth]{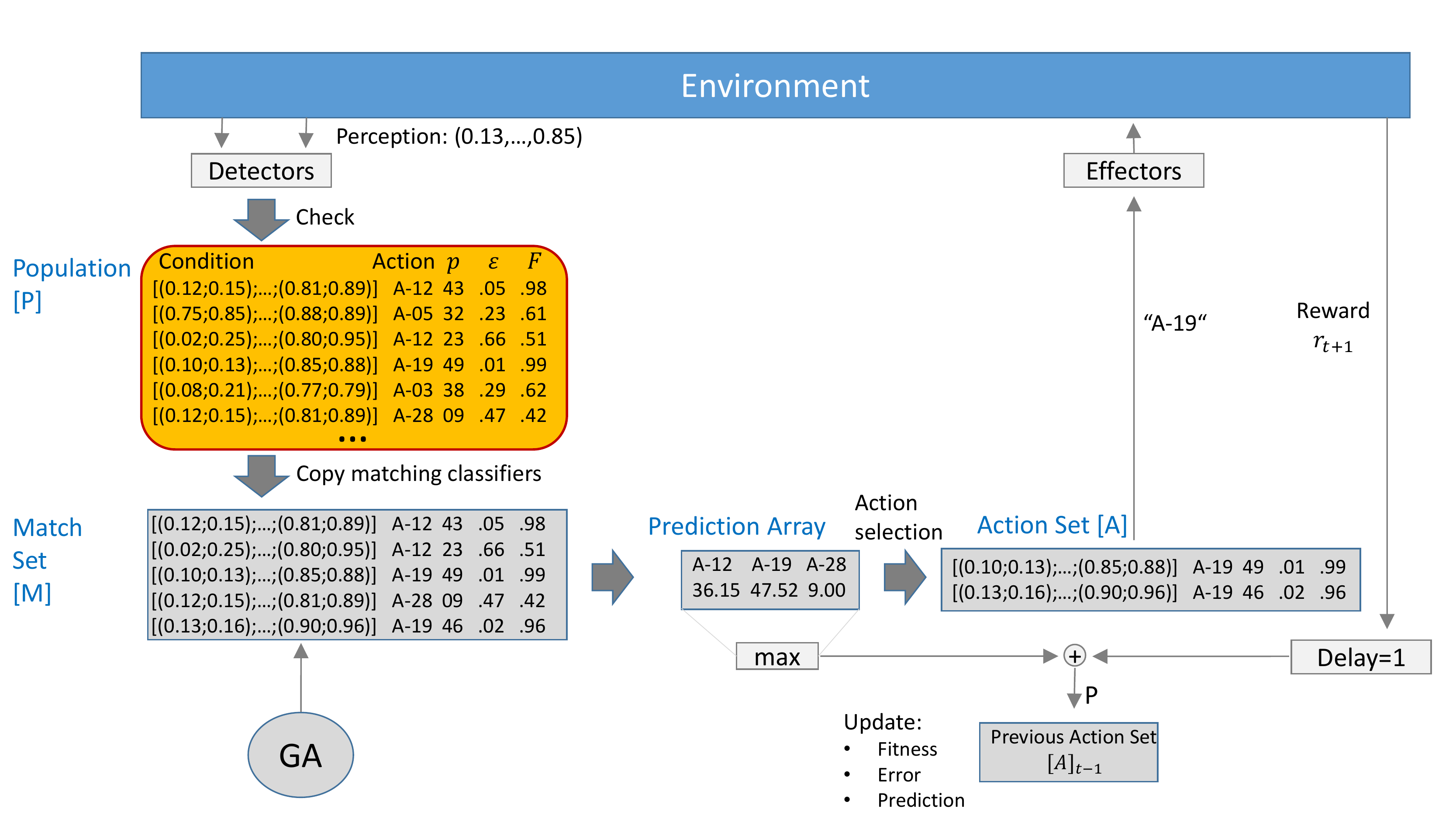}
\caption{Schematic illustration of XCS-R with example population.}
\label{fig:XCS}
\end{figure*}

Following these requirements, the class of ``Learning Classifier Systems'' (LCS) \cite{Wilson2000} in general and the variant ``Extended Classifier System'' (XCS) by Wilson \cite{Wilson1995} in particular have proven to provide suitable techniques. An alternative is the usage of Deep Reinforcmeent Learning technology that suffers from the missing interpretability (i.e., this is typically neglected in SASO systems since te user or administrator has no insights in what the system will do and why). An overview of LCS variants can be found in \cite{sigaud2007learning}. In contrast, the konlwedge of LCS is stored in an explicit rule-base that is evolved over time. LCS/XCS are flexible, evolutionary rule-based RL systems---we rely on the XCS-R variant ('R' for real values) in the following, since this is the variant used in intelligent/SASO systems \cite{DAngeloGGGNPT19}. XCS-R relies on a population $[P]$ of classifiers, where each of these classifiers is an ``IF-THEN'' rule with evaluation criteria that partition the input space $X$ and thus approximate the problem space locally. A steady-state niche genetic algorithm (GA) is responsible for improving the coverage in the match set $[M]$ (that contains all classifiers where the condition part matches the current conditions). Based on the set of matching classifiers, a fitness-weighted prediction is calculated for all actions proposed by classifiers contained in $[M]$---using these values, a roulette-wheel approach is followed to select the action to be applied. All classifiers in $[M]$ promoting the selected action are transferred to the action set $[A]$ for later consideration. For learning purposes, a discount is taken from the classifiers contained in $[M]$, a (delayed) reward is considered, and the classifiers of the last action set (i.e., those being responsible for the last behaviour) are updated accordingly (i.e., using a learning rate $\alpha$). Figure~\ref{fig:XCS} illustrates the process for the variant XCS. 

\begin{figure*}[ht!]
\centering
\includegraphics[width=0.85\textwidth]{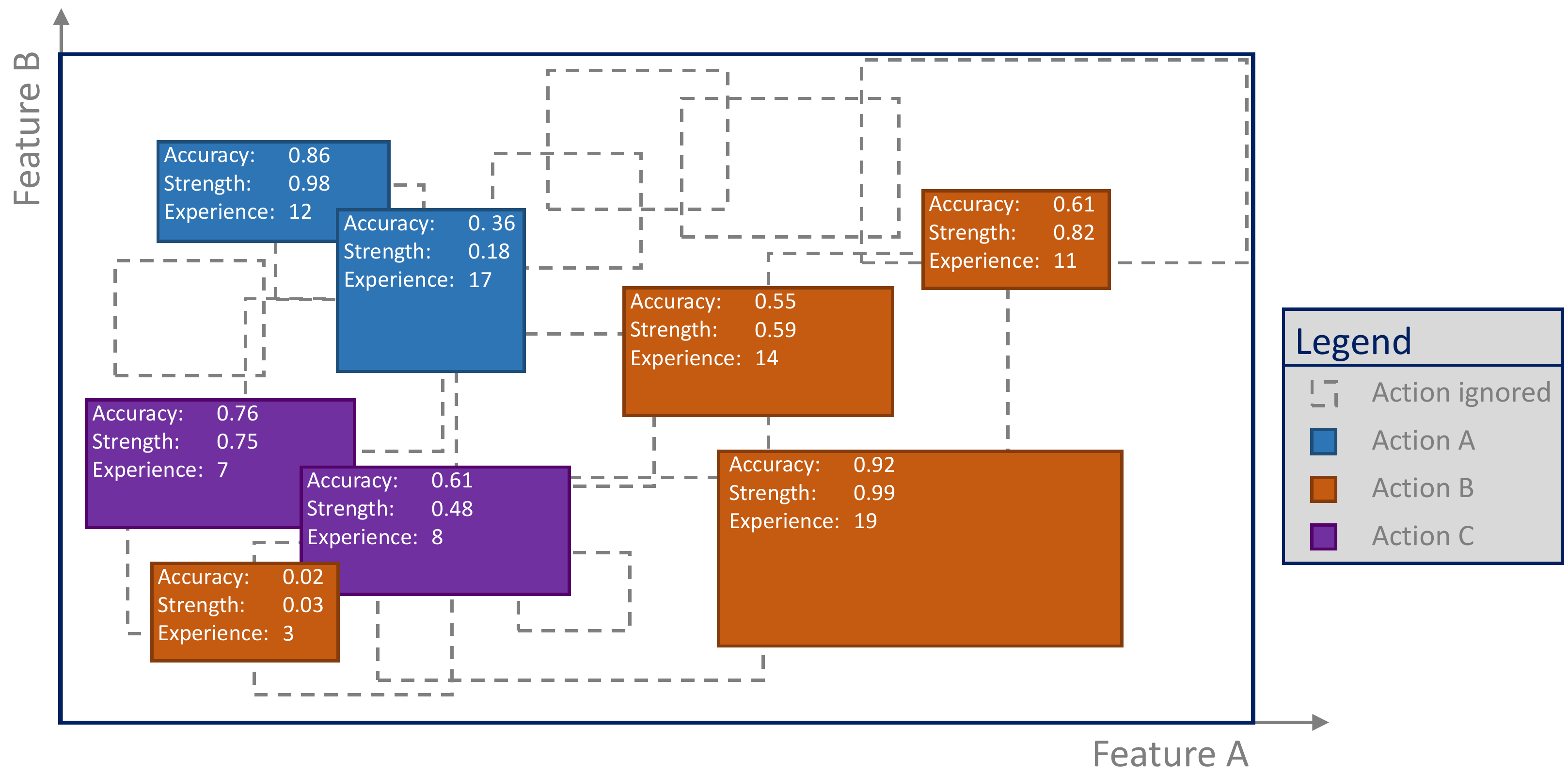}
\caption{Example for a population of XCS covering the condition space.}
\label{fig:RL}
\end{figure*}

The classifiers are usually initialised with partially predefined initial but also randomly selected values. A covering mechanism generates new classifiers if $[M]$ contains too few or inappropriate classifiers. However, this is a rather reactive behaviour. As mentioned before, a steady-state niche GA is constantly creating new knowledge in form of classifiers: Successful classifiers are selected to be reproduced, recombined, and mutated to search the local neighbourhood of the environmental niche with the result of a better coverage. In particular, this means that a globally optimal subregion of the input space is evolved that is characterised by a high accuracy in predicting the corresponding payoff (or reward). Again, this GA can be considered to act re-actively since it is performed in cycles and only considers 'good' classifiers without any kind of creativity. Figure~\ref{fig:RL} illustrates an example of a population of XCS in a two-dimensional search problem: The space is unevenly covered by classifiers with different size of the condition part and different evaluation values. In general, XCS keeps track of the three primary attributes $p$ (prediction of the expected payoff), $\epsilon$ (prediction error observed over time), and $F$ ('fitness' of the classifier) and several further 'book keeping' attributes such as the experience (number of occurrences in the action set), numerosity (number of contained sub-classifiers after subsumption), etc.

Although XCS has proven to be applicable to real-world problems (e.g., \cite{TB+11}), it comes with some major drawbacks:
\begin{enumerate}
    \item Exploration vs. exploitation is done by a roulette-wheel approach considering the $p$ and $F$ values of classifiers in $[M]$ only. There is no controlled exploration behaviour and no adaptation of the learning rate $\alpha$.
    \item XCS builds up and manages its population in a purely reactive manner: It generates and adapts classifiers only in those parts of the input space where it observes stimuli. It does not generate classifiers for other niches pro-actively and, hence, is not prepared for new conditions.
    \item XCS generates new classifiers either by genetic operations in a niche (i.e., the GA) or randomly (the covering mechanism). It does not use existing knowledge of the entire population or relies on a goal-oriented mechanism for the covering part.
    \item The number of classifiers contained in the population is limited. Consequently, XCS contains mechanisms to aggregate classifiers into one and to delete less promising classifiers. However, it does not actively manage its population.
    \item The population just stores the best actions following the current experiences. However, it does not make use of knowledge about unsuccessful attempts.
    \item The approach can handle concept drifts in a purely passive manner, but it does not contain mechanisms to detect this at runtime and find efficient reactions.
\end{enumerate}

The idea of this article is to develop the concept for a novel XCS variant that is able to handle these issues. In particular, this means to turn the passive or reactive XCS into an active or proactively acting system, to which we refer as active reinforcement learning. Therefore, we make use of concepts from the active learning \cite{settles2009active,settles2012active} domain---which is introduced in the next paragraphs.

\subsection{Active Learning}
\label{sec:bg:al}

Active Learning (AL) is a semi-supervised learning paradigm that is based on managing the sample selection process in the sense that it selects the most informative samples for labelling. The goal is to achieve a high performance (e.g., classification accuracy) with as few samples as possible. Application areas where AL has been successfully applied include: drug  design \cite{kangas2014efficient}, text classification \cite{chu2011unbiased}, or malicious software detection \cite{nissim2014novel}. The term AL has initially been defined by Settles as follows: ''“Active learning systems attempt to overcome the labelling bottleneck by asking queries in the form of unlabelled instances to be labelled by an oracle. In this way, the active learner aims to achieve high accuracy using as few labelled instances as possible, thereby minimising the cost of obtaining labelled data.'' \cite{settles2009active}.

In general, we distinguish between three major AL paradigms: stream-based active learning (SAL) \cite{atlas1990training}, membership query learning (MQL) \cite{angluin1988queries}, and pool-based active learning (PAL) \cite{lewis1994sequential}. The SAL paradigm assumes a stream of data, i.e., the data points show up one after the other and the learner has to decide either to 'buy' (i.e., query an oracle) a label or not. In particular, this means if the learner decides not to buy a label for a specific data point, he will not be able to access that data point again. 

In the MQL paradigm, the learner works independently from a stream of incoming samples and can possibly request labels for any unlabelled instance in the input space. In particular, this means to allow for queries that the learner generates de novo.

\begin{figure*}[ht!]
\centering
\includegraphics[width=0.80\textwidth]{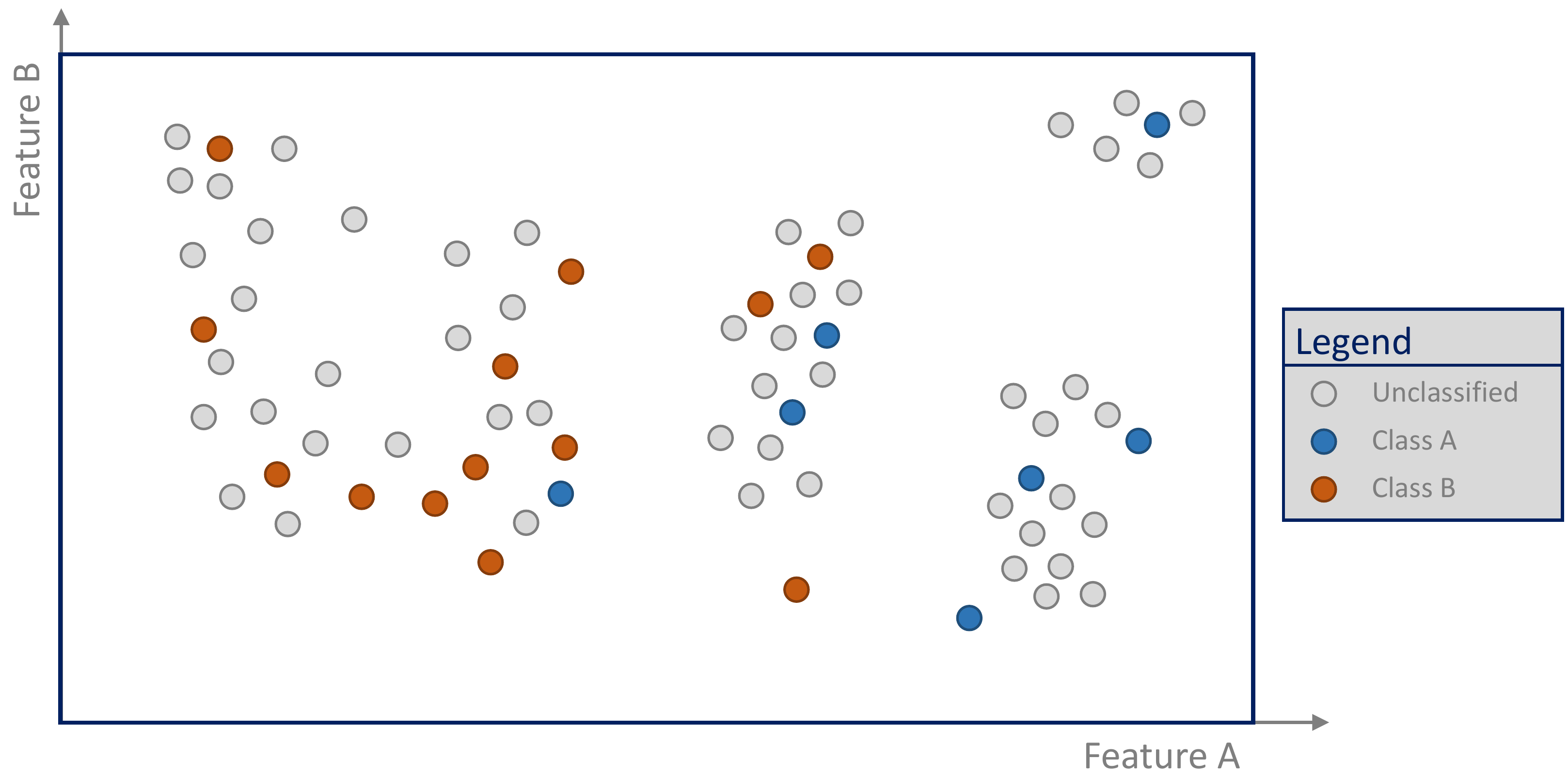}
\caption{Example of labelled and unlabelled samples in a pool-based AL setting.}
\label{fig:AL}
\end{figure*}

In contrast, the standard PAL process starts with a large pool of unlabelled samples and a small set of labelled samples. Figure~\ref{fig:AL} illustrates such a pool with already labelled samples (blue and orange) and unlabelled samples (grey). The training process is organised in cycles and consists of the following steps: (1) A model such as a classifier is trained using the already labelled samples, (2) The selection strategy of the active learner identifies a query set (the next samples to be labelled) from  the pool of unlabelled samples, (3) The selected samples are presented to the oracle (e.g., a human), which provides label information, and (4) The knowledge model is updated. Finally, (5) the process terminates if a stopping condition is met, otherwise the next learning cycle is started.

PAL heavily relies on the particular implementation of the selection strategy. Hence, several alternatives are available in literature---the most prominent examples include the following:
\begin{itemize}
    \item \textit{Random selection}, also called passive sampling: Selects instances randomly for labelling, which means that it is free of heuristics and parameters.
    \item \textit{Uncertainty sampling:} Select those instances where the learner are least certain about the label, with different measures being used for quantification of this uncertainty (e.g., posterior, margin, or entropy) \cite{atlas1990training}.
    \item \textit{Ensemble-based strategy:} The basic idea here is to train different base classifiers by using different subsets of the training data and choose the next sample to be queried by identifying the strongest disagreement between these base classifiers \cite{seung1992query}.
    \item \textit{Expected Error Reduction}: Estimates the generalisation error of a classification model given a validation set, which means that no labels are required \cite{RMC01}.
    \item \textit{Density weighted Uncertainty Sampling (DWUS)} An uncertainty score is weighted with a candidate's density. Thereby, outlier are unlikely to be considered for labelling \cite{donmez2007dual}.
\end{itemize}

In general, all these strategies aim at assessing the current knowledge of the classifier and, based on this assessment, determine the most informative samples. In the following section, we describe how this basic idea in combination with the different approaches of these basic selection strategies can be used to turn purely reactive RL systems into actively learning RL systems.

\section{\uppercase{Active Reinforcement Learning}}
\label{sec:approach}

\begin{figure}[ht!]
\centering
\includegraphics[width=0.99\columnwidth]{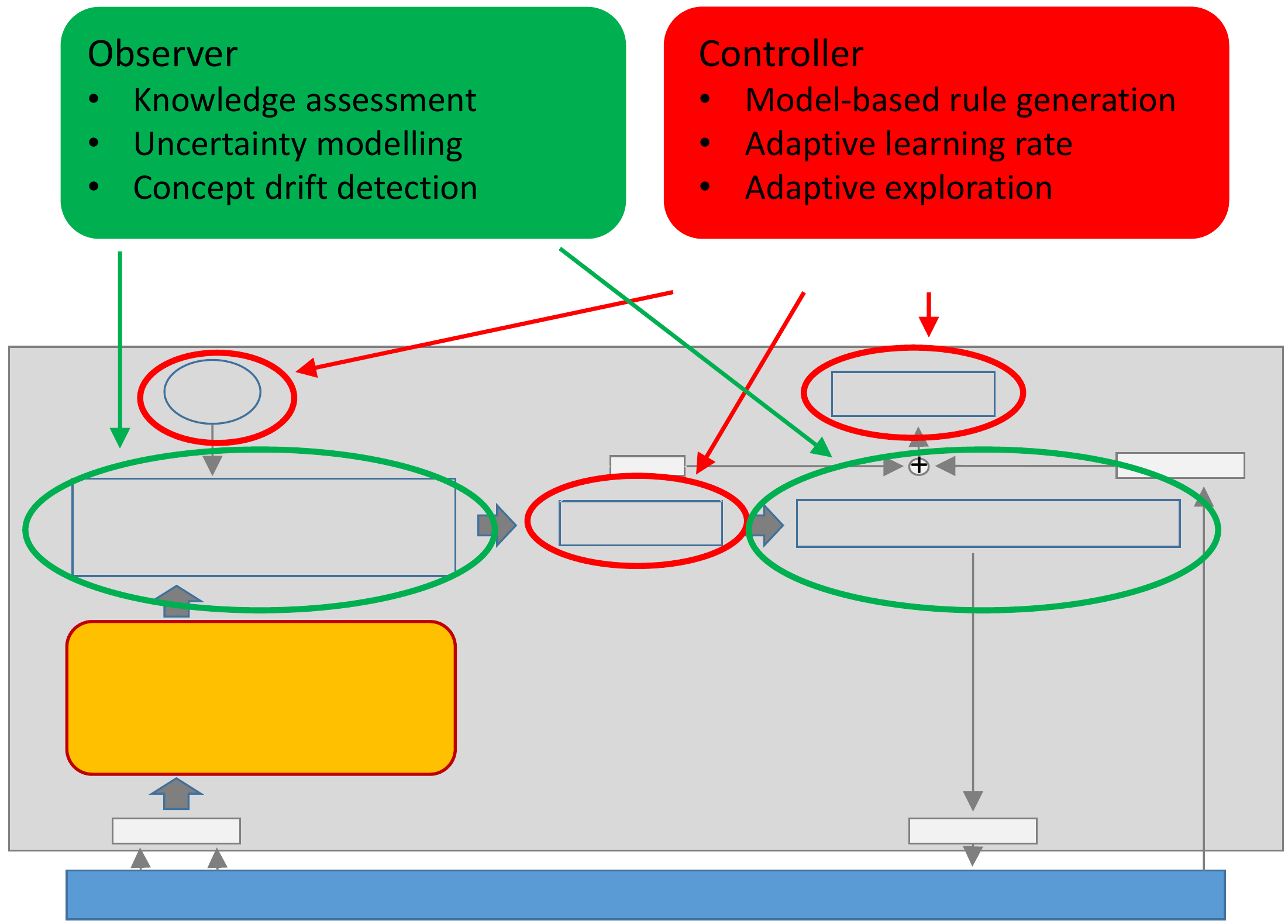}
\caption{Self-adaptive XCS with observer/controller tasks.}
\label{fig:obsCon}
\end{figure}

\noindent We use the term 'active reinforcement learning' (ARL) for RL systems that are self-aware of their knowledge and adapt their own learning behaviour accordingly. This is in contrast to the previously seen purely reactive and passive strategies typically followed by RL systems (see Section~\ref{sec:bg:rl}. Figure~\ref{fig:obsCon} illustrates an abstract view of an XCS presented in Figure~\ref{fig:XCS} and adds an adaptation mechanism on-top of it that follows the design concept of Organic Computing \cite{TP+11}. In general, active RL comprises two different aspects that need to be considered subsequently: i) establishing self-awareness of the own knowledge and ii) active control of knowledge discovery parts of the RL system. This corresponds to the observer and controller parts of the design concept. We summarise the corresponding tasks in the following paragraphs.

\subsection{Aspects of ARL}
\label{sec:app:aspects}

\textbf{Observation tasks:} As basis for active management of knowledge and guided reinforcement learning behaviour, the system needs to become aware of its own, currently existing knowledge. This knowledge is stored in the population (i.e., the available classifiers as illustrated by Figure~\ref{fig:RL}): in the entirety of the search space covered by condition parts, in the diversity of actions, and in the different evaluation criteria (i.e., prediction, error, fitness---but also experience, numerosity, etc.). In order to allow for a kind of self-awareness of the existing knowledge, an active reinforcement learning systems needs the following capabilities:
\begin{itemize}
    \item Assessing the quality (i.e., the accuracy of predictions of rewards/feedback in combination with the strength of the expected reward) of existing knowledge as distribution model for the entire search space: This allows for identifying regions that have insufficient knowledge (which then needs to be addressed actively by the controller part).
    \item Modelling the uncertainty of the existing knowledge: Assessing the certainty assigned to the different regions of the search space can be done using the aspect 'experience' per classifier aggregated over the different classifiers covering a niche. This allows for determining a score for each possible niche that---in relation to other niches--- provides the basis for deciding if this niche needs to be explored in more detail.
    \item Assessing the 'appropriateness' of the knowledge for the next steps: An RL system typically traverses the search space in terms of a trajectory. This means that there is a local dependency between the current situation and the next situation, since this is the result of the chosen action and environmental influences. In order to act actively, the RL system needs to make use of these trajectories and predict the set of next situations (probably probabilistically to quantify the uncertainty assigned to these predictions). Based on this, it can assess in advance if appropriate classifiers will be available for the corresponding match and action sets. Otherwise, alternatives need to be generated.
    \item Detecting meta-features of the learning problem: Typically, RL assumes that the underlying learning problem is static, i.e., challenges such as concept drift and concept shift do not occur. Alternatively, it is assumed that concept drift is so slow that the evolution of the population automatically addresses this problem. However, intelligent systems have to operate under real-world conditions and based on sensor/actuator constellations, which may result in challenges such as i) integration of novel components, ii) hardware characteristics such as wear, or iii) any kind of novel environmental processes. Consequently, the systems needs possibilities to detect such changes in the underlying problem domain, possibly resulting in re-consideration of already learned knowledge.
\end{itemize}

\textbf{Control tasks:} Taking the different mechanisms of the observer part as input, the RL system has to adapt its own learning behaviour accordingly. In general, this means to manage parameters such as learning speed but also to generate promising classifiers for inappropriately covered niches of the search space. To allow for a kind of self-management of the learning behaviour, an ARL systems needs the following capabilities:
\begin{itemize}
    \item Model-based classifier generation: Instead of randomly generating new classifiers or applying a genetic algorithm to the action or match set, the RL system needs more sophisticated techniques for classifier generation that make use of the available knowledge.
    \item Adaptive exploration rate: Currently, XCS decides about the next action based on the information contained in the prediction array by applying a roulette-wheel approach. This gives a higher probability to better performing actions (i.e., exploitation) and lower probability to worse performing actions (i.e., exploration). However, such a static assignment of probabilities based on the niche is a good idea at startup, it result in undesired exploration when the population is already converged. Consequently, there are different strategies available in literature that aim at controlling the learning rate. An ideal strategy comes with a high learning rate when the population is not appropriate (e.g., startup or after a concept drift/shift) and with no exploration if the population converged to the optimum. Consequently, the RL system needs a mechanism to assess the current exploration needs and to control the exploration probability accordingly.
    \item Adaptive learning rate: The selection probability is just one means to parameterise the learning behaviour in XCS. A second aspect is given by the learning rate that is responsible for controlling to which degree a classifier's evaluation criteria are shifted into the direction of the actually received reward/feedback signal. Here, the same argumentation holds as above, resulting in the need to control the learning rate dynamically at runtime.
    \item Population management: The size of the population is limited in XCS to allow for a 'selection pressure', i.e., less performing classifiers are continuously replaced by novel candidates. The result of this process is that, after convergence, the population is (theoretically) perfectly optimised to the underlying problem and only stores the most promising responses for occurring situations. However, in the presence of permanent change, deleting information about sub-optimal or even bad behaviour (and consequently the encoding classifiers) is highly inefficient. Reasons include: i) A concept drift may render deleted classifiers necessary again, ii) generating new classifiers efficiently means to avoid bad behaviour, etc. As a counter measure, the RL system needs a mechanism for active management of the current population which makes, e.g., use of a backup memory that allows for 'constructive forgetting' and 'remembering'.
\end{itemize}

There are further highly important aspects when turning XCS into an ARL system, which need to be taken into consideration:
\begin{itemize}
    \item XCS is just one RL paradigm, although probably the most prominent in operating intelligent systems. However, the insights gained in developing an active XCS need to be transferred to other RL paradigms for generalisation purposes.
    \item An intelligent system seldom operates alone. In turn, use cases often contain several similar systems operating in a shared environment. Consequently, the mechanisms identified before can make use of even more existing knowledge (i.e., the populations of other systems of the same kind) via communication. This, however, may increase the uncertainty related to the knowledge information since the underlying characteristics (such as the sensor equipment) may not be identical.
\end{itemize}


\subsection{State of the art in active reinforcement learning}
\label{sec_app:sota}

\textbf{Assessment of the existing knowledge:} Stein et al.\ defined so-called 'knowledge gaps' as parts of the input space (niches) that are not sufficiently covered by classifiers \cite{SteinTDHM18}. This means that the niche is either not covered at all or only covered by classifiers with low performance values. The authors propose to derive an abstract representation of the classifier distribution that is used to identify niches where knowledge is needed. In \cite{stein2017toward}, a basic combination with approaches from the field of AL has been derived that should make use of concepts such as query synthesis, uncertainty sampling, and query by committee---however, this just provides the idea and can be seen as preliminary work to this contribution.

\textbf{Considering external knowledge in LCS}: Especially for Michgan-style LCS, Urbanowicz et al.\ presented a concept to integrate the knowledge of external experts to establish a guided discovery process of novel classifiers \cite{urbanowicz2012using,urbanowicz2015exstracs}. In general, the approach is to make use of a set of hand-crafted heuristics (i.e., 'expert knowledge') to improve the generation of novel classifiers in comparison to the random-based strategy typically used. This means that the probabilities of actions covered in situations of classifiers under construction are pre-defined (i.e., derived in an offline pre-processing step) and not adaptive to the underlying problem.

As an alternative, Najar et al.\ presented a learning model where a human 'teacher' guides the RL system (a robot) by using teaching signals (i.e., showing what to do) \cite{najar2015social,najar2015socially}. The idea is to allow the learner to imitate the behaviour of the teacher and therefore improve its learning efficiency. However, the approach is limited to cases where learner and teacher can explicitly do the same task, where a (human) teacher is available during the entire learning process, and where knowledge is already available by teachers.

In the context of Organic Computing \cite{MST17}, a safety-based generation of novel classifiers has been presented \cite{TB+11}. Here, the LCS-variant XCS is not allowed to generate new behaviour. In contrast, it operates on the existing classifiers only. The GA responsible for discovering appropriate actions in specific situations has been removed from the online component and coupled with a simulation of the underlying intelligent system. As a result, only tested behaviour is added to the population. To remain operable (i.e., find answers in case of missing knowledge), the covering component copies the nearest classifier (i.e., based on the Euclidean distance defined in the search space of the condition parts), widens its condition part to fit the new situation, and uses this as default classifier \cite{PT+11}. However, the approach requires the existence of a 'digital twin' during the discovery process.

Also in the context of OC, Stein et al.\ proposed to make use of interpolation techniques to generate new classifiers \cite{SteinRTH16}. The basic idea is that the best action is probably a compromise of the actions proposed by the best surrounding classifiers in the niche. Consequently, novel classifiers are generated not by using a simulation-coupled GA as in \cite{TB+11,PT+11} but by interpolation of the neighbours---which is much faster. The approach has been further extended to alternative usage patterns of interpolation \cite{SRTH17} and a concept for proactive knowledge construction that, however, has not been realised yet \cite{stein2017toward}. Still, this assumes a gradual change in the action rather than allowing for shifts in the action distributions.

Following a more generic approach, Nakata et al.\ developed an example of a weighted complete action map, which is intended to evolve a population that is complete in terms of learning the entire situation-to-action-mapping. In particular, this means to assign more classifiers to the highest-return actions for each state, for instance \cite{nakata2015should}. This continues work by Kovac\ (see e.g. \cite{kovacs1999strength}, where he investigates the trade-off between strength and accuracy and consequently emphasises issues related to the change in fitness.

\textbf{Management of the population:} In \cite{butz2011xcsf}, Butz and Sigaud proposed a different approach to classifier deletion. They handled the selection problem (i.e., identification of classifiers to be deleted to keep the population size below a given threshold) locally rather than globally as before. This is intended to avoid deleting classifiers in niches that can be considered as knowledge gaps using the wording of Stein et al. from \cite{SteinTDHM18}. However, this does not tackle the problem of actively managing the population.

In XCS, a subsumption mechanism combines similar rules, and a randomised deletion mechanism removes classifiers of a low fitness from the population. In \cite{FredivianusPS10}, the discovery component is altered by introducing a modified rule combining technique. The goal is to create maximally general classifiers that match as many inputs as possible while still being exact in their predictions. The approach considers previously learnt knowledge and infers generalised classifiers from the existing population. It has been extended to real-valued instances of XCS in \cite{FredivianusKS12}. Despite providing a heuristic for rule combining, the approach covers just one aspect of the management problem.

\textbf{Imbalanced data:} RL systems in intelligent systems learn in a reactive manner by considering the current environmental conditions as input. In other words, the different input situations the RL system has to react on are highly imbalanced, with some regions of the search space being never or only extremely seldom covered. The learning mechanism of XCS under imbalanced data has been investigated and theoretically modelled by Orriols-Puig et al.\ in \cite{orriols2009evolutionary}. The authors focus on rare cases and react to this challenge by adapting XCS' parameters (i.e., the learning rate and the threshold for GA activation). However, this does not explicitly address the problem of identifying limited knowledge or steering the exploration---but it provides a basis for techniques to modify XCS parameters at runtime.

\textbf{Novelty search:} LCS are part of the overall field of evolutionary algorithms (EAs). Here, approaches such as the 'Novelty Search Algorithm' \cite{lehman2008exploiting} by Lehman and Stanley have been presented that exchange the traditional evolution process based on a fitness function by one that puts an emphasis on searching for novel behaviours. This may be beneficial if the search space of possible actions is unknown and consequently can provide an element for action discovery in active RL systems.

\textbf{Exploration strategies:} Initially, RL systems used static exploration strategies with the $\epsilon$-greedy strategy as probably most popular one \cite{SuttonB1998}. Here, a certain fraction of attempts ($\epsilon$) are dedicated to exploration, while the remaining $1-\epsilon$ fraction is used for exploiting the currently best known action. Variants of static exploration schemes include a random strategy (also called random walk, i.e., selecting randomly which action to use), or a softmax strategy (i.e., an $\epsilon$-greedy approach with modified selection probabilities to better reflect the currently learned reinforcement values) \cite{SuttonB1998}.

In contrast, the idea of the 'decreasing $\epsilon$-strategy' \cite{vermorel2005multi} is to have a higher probability for an explorative action selection at the beginning and to decrease this towards the end of the learning process. In this way, an attempt is made to ensure appropriate exploration of the search space before the agent finally makes primarily exploitative action selections. The limitations of this approach include the static decreasing behaviour (i.e., no real adaptiveness) and the characteristics of real-world problems, since there are typically no final states. A similar alternative is the '$\epsilon$ first' strategy \cite{vermorel2005multi}--here, only two different values for $\epsilon$ are considered: a high one (e.g., 1) at the beginning and a low one (e.g., 0 or close to 0) after sufficient exploration. The limitations are similar as before but further emphasise the need to define 'sufficient' exploration. 

A third category of exploration strategies is defined by the 'meta-softmax-strategy' \cite{schweighofer2003meta}. The idea is to adapt the learning rate, the discount factor and the exploration probability dynamically in response to the difference of two values calculated from the feedback signal: a 'mid-term reward' (i.e., considering a few rewards) and a 'long-term reward' (i.e., considering the rewards of a longer window). For both, the averaged rewards are determined for the considered window and if the difference is above a pre-defined threshold, the exploration criteria (i.e., learning rate, discount factor, and exploration probability) are increased. This is already adaptive in the sense of our notion of ARL, but it still follows a reactive mechanism rather than a well-defined active decision. Closely related to the 'meta softmax' strategy is the class of 'value-difference-based-exploration' strategies \cite{tokic2010adaptive}. Here, the exploration parameters of the learning strategy are determined based on the difference of the 'values' of the individual actions. In particular, this value difference denotes the product of the learning rate and the temporal difference error. It can not be mapped directly to the evaluation criteria of XCS, since there is no individual 'value' of a classifier. However, the prediction error in XCS can be understood as a corresponding indicator. We have to consider differences between expected reward/feedback and observation as one aspect of the strategy to decide about adaptations of the explorative parameters.

\section{\uppercase{Research Roadmap}}
\label{sec:roadmap}

\noindent To realise ARL based on XCS, different challenges need to be addressed. In this section, we outline a corresponding research roadmap. Therefore, we define the most urgent challenges that contribute to the aspects of ARL as introduced previously. We organise these challenges in terms of observer, controller, or overall system tasks. For each challenge, we initially discuss the goal and subsequently propose a concept how this challenge can be addressed using. Thereby, we either make use of concepts from the state of the art or we put an emphasis on possible strategies to to close the gap in research.

\subsection{Observer-related challenges}
\label{sec:rr:obs}

The goal of the observer part is to establish a self-assessment of the existing knowledge and a subsequent prediction of possible next states. This mainly comprises the challenges assessment of the population, prediction of the next states, and modelling the entire learning problem.

\textbf{Challenge 1 -- Assessment of the existing population:} The goal of the first challenge is to establish a continuous self-assessment of the existing population. This mainly refers to the task of identifying less covered niches or niches with inappropriate knowledge. 

The different AL strategies as outlined in Section~\ref{sec:bg:al} are already fulfilling the task of assessing the input space. However, the major difference is that the population is a set of hyper-rectangles covering the search space (see Figure~~\ref{fig:RL}) rather than the individual points in the AL approach (i.e., the samples in Figure~\ref{fig:AL}). This means that we have to turn the sample-distribution into a continuous distribution, e.g. by using kernel density estimation techniques. However, the hyper-rectangles defining the condition parts of the classifiers are combined with a kind of label information (i.e., actions), but they also consider uncertainty values (i.e., the evaluation criteria such as accuracy, strength, experience or numerosity). The assessment must be able to derive values for these aspects as well. Further, an ordering of knowledge gaps (i.e. a ranking) will most certainly be based on several criteria aggregated to an individual score. Here, aspects such as described for the 4DS strategy in AL can play a major role, see \cite{reitmaier2013let}.

\textbf{Challenge 2 -- Prediction of the next match sets:}
The goal of the second challenge is to predict the next observations to check whether the match set will be suitable or not. This can then serve as basis to proactively generate new classifiers.

A possible approach is to model the sequence of situations as a sliding window approach. Within this window, a trajectory can be determined --- which is then used to predict possible next situations. To further improve the predictions, this has to take aspects such as velocity and probability of occurrence into account. In addition, the prediction may become subject to runtime learning as well by using state-of-the-art techniques. Based on this, the match set can be generated and an analysis of the contained classifiers can be performed: Is there enough diversity? Is the overall fitness-weighted prediction above a certain threshold? Is the niche sufficiently covered?

\textbf{Challenge 3 -- Learning problem modelling:}
The goal of the third challenge is to generalise the first challenge by explicitly modelling the problem space next to the knowledge stored in the population.

A possible approach lies in the integration of an additional, continuously-defined representation of the learning problem that is update based on any received feedback. This can be done, for instance, by training a neural network taking the situation as input and providing the most appropriate action as output (or estimating the payoff for a situation-action pair). Such a solution -- and therefore the reason why we favour the XCS-based approach -- suffers from the representation being not interpretable by humans and not explaining the behaviour. As an additional knowledge base, this may serve as input for more efficient and reliable classifier generation and management.

%
%
\subsection{Controller-related challenges}
\label{sec:rr:con}

The goal of the controller part is to increase the degree of autonomy of (reinforcement-based) learning in intelligent technical systems. This actually turns the RL system into an ARL system by making use of the information provided by the observer part---and it mainly comprises the challenges population management, controlled classifier generation, adaptive exploration probability, adaptive learning rate, and control of the adaptation speed.

\textbf{Challenge 4 -- Population management:} The goal of population management is to turn the reactive replacement approach with a defined maximum number of classifiers in the population into a proactive management.

A possible approach is to introduce external offline memories that serve as reservoir for removed classifiers. The knowledge encoded in these classifiers can also be used as reference when generating new ones. Especially if a change in the structure of the learning problem is noticed, the system can check if it could switch back to previously removed classifiers, which would perfectly cover oscillating behaviour, for instance. Such a mechanism needs to be augmented with a suitable selection scheme as well as a mechanism that ideally abstracts from the individual classifiers. Thereby, not only the memorisation of older classifiers plays an important role, but also techniques to realise 'constructive forgetting', i.e. explicitly select knowledge to be removed to allow for novel behaviour (which not necessarily refers to 'bad' classifiers as currently).

\textbf{Challenge 5 -- Controlled generation of new rules:}
The goal of this aspect is to proactively generate new classifiers.

There are already a few attempts in literature: \cite{TB+11} uses a simulator to generate new classifiers and \cite{stein2017toward} describes interpolation-based ideas. Following these ideas, new classifiers can be generated either using side-knowledge (i.e., the external neural network mentioned above), simulation, external memories of replaced classifiers or the neighboured classifiers. The scope can be extended towards incorporation of other opportunistically available knowledge sources as explained in \cite{CalmaKST17}. This is closely related to the concept of curiosity, which is discussed in detail in \cite{wu2013curiosity}.


\textbf{Challenge 6 -- Active configuration of exploration probability:}
The goal of this challenge is to replace the roulette-wheel exploration approach by an adaptive mechanism.

As outlined in Section~\ref{sec_app:sota}, there are several adaptive exploration schemes available in literature. They can serve as starting point for altering the decision logic of XCS as well. However, this needs to be combined with change detection techniques or novelty detection techniques (such as \cite{GruhlST21} taking, e.g., the trajetories through the search space as input) to identify conditions where the reinforcement behaviour is differing from previous and expected behaviour.

\textbf{Challenge 7 -- Adaptive control of the learning rate:}
The goal of this aspect is to replace the static definition of the learning rate by adaptive techniques.

Similar to the challenge before, cases where the underlying learning problem is changing require a faster adaptation of the existing knowledge. In turn, already settled knowledge bases require smaller or no update at all to avoid oscillations due to noise feedback. A solution can adapt the reinforcement rate in a way that $\alpha$ is increased in case of changing conditions (e.g. detected drift). Alternatively, the value of $\alpha$ depends on the niche of the search space (number of classifiers, fitness, numerosity, etc) and can be defined per-niche rather then globally.

\textbf{Challenge 8 -- Adaptation speed:} The goal of this aspect is to replace the static cycle-based activation scheme of the learning technique with an adaptive variant.

Currently, the RL loop is performed in given cycles of fixed duration. However, the frequency of adaptation performed by the intelligent system is not necessarily desirable to be constant. For instance, in traffic control the conditions are almost constant during the night, resulting in seldom adaptation needs. On the other hand, rush hour handling would benefit from even shorter adaptation cycles. This maps also to the idea that the learner is able to dynamically change the number of observations until it decides to adapt. Consequently, an approach could rely on assessing the stability of the observed condition and -- based on such a score -- adapt the frequency of adaptation. However, this has to avoid self-lock-in problems, for instance.

\subsection{System-related challenges}
\label{sec:rr:sys}

Besides the observer and controller tasks, there are system-wide challenges for ARL. This mainly comprises the challenges multi-dimensional feedback signals, collaborative awareness, and indirect feedback.

\textbf{Challenge 9 -- Active reward requests:}
The goal of this aspect is to integrate human users as additional knowledge source.

In accordance with the basic idea of AL, we introduce the user or administrator of the system as additional knowledge source. For unknown conditions or alternative classifiers, the system may query the user for a feedback signal that is not stemming from the observed environment. This means to fully integrate the self-assessment of the knowledge distribution combined with the uncertainty estimation as known in AL and incorporate the corresponding selection strategies, possibly with a given budget and only if a user is available. This entails the need for a availability models for the user, maybe even augmented with (learned) models of their expertise.
	
\textbf{Challenge 10 -- Multi-dimensional feedback signals:} The goal of this aspect is to replace the reward signal by a vector comprising several utility functions at the same time.

A typical intelligent system has to tackle more than one goal at the same time. Integrating the different utility values into one score is always a trade-off and looses information. Consequently, the algorithmic logic needs to be adapted in a way that the single value is replaced by a vector representation. This has impact on all stages of the learning process. However, there is a first solution available \cite{BeckerHT12} that serves as a basis for tackling the problem.

\textbf{Challenge 11 -- Indirect feedback signals and feedback with uncertainty:} The goal of the last aspect is to combine the explicit with possible further implicit feedback signals.

A possible approach establishes a mechanism that tries to determine additional feedback signals for consideration. An intuitive approach relies on using neighbouring systems of the same kind that act in a shared environment: Thy can provide (negative) feedback about adaptation decisions that have a negative impact on the utility of the neighbours. This would turn the methodology as defined in \cite{RudolphTH19} into a direct representation. However, this also implies that feedback signals have to be considered in different ways, e.g. based on an uncertainty value or the type of feedback (indirect vs. direct). 

\section{\uppercase{Conclusion}}
\label{sec:conclusion}

\noindent This paper discussed the limitations of current techniques applied to the self-adaptation task in intelligent systems. The major observation is that currently the specific learning parts are processed in an isolated manner. In particular, the main learning technique is typically a reinforcement learner that learns in a purely reactive manner. On the other hand, machine learning paradigms such as anomaly/novelty detection or active learning are able to provide a better self-awareness of the underlying observed behaviour and processes.

Based on this observation, we proposed a concept to integrate current sophisticated reinforcement learning techniques --- in particular, the class of Learning Classifier Systems --- with concepts from the other domain. The goal is to establish an integrated approach, which we called 'active reinforcement learning'. The major advantage of such a technique lies in the 'proactiveness', i.e. the possibility to act self-determined (optimised, planned) rather than purely reactive.

We used the basic design pattern from the domain of Organic Computing, i.e. the Observer/Controller pattern, as a reference model for intelligent systems. Based on the constituent parts, we derived a research roadmap towards closing the gap to an active reinforcement learning system. This resulted in the definition of nine challenges. However, this list does not claim to be exhaustive, but reflects the most urgent steps towards an ARL approach. In our current and future work, we focus on these challenges.

\bibliographystyle{apalike}
{\small
\bibliography{references}}

\begin{thebibliography}{}

\bibitem[Angluin, 1988]{angluin1988queries}
Angluin, D. (1988).
\newblock Queries and concept learning.
\newblock {\em Machine learning}, 2(4):319--342.

\bibitem[Atlas et~al., 1990]{atlas1990training}
Atlas, L.~E., Cohn, D.~A., and Ladner, R.~E. (1990).
\newblock Training connectionist networks with queries and selective sampling.
\newblock In {\em Advances in neural information processing systems}, pages
  566--573.

\bibitem[Becker et~al., 2012]{BeckerHT12}
Becker, C., H{\"{a}}hner, J., and Tomforde, S. (2012).
\newblock Flexibility in organic systems - remarks on mechanisms for adapting
  system goals at runtime.
\newblock In {\em Proc. of 9th Int. Conf. on Inf. in Control, Automation and
  Robotics}, pages 287--292.

\bibitem[Butz and Sigaud, 2011]{butz2011xcsf}
Butz, M.~V. and Sigaud, O. (2011).
\newblock Xcsf with local deletion: preventing detrimental forgetting.
\newblock In {\em Proc. of 13th An. Conf. Companion on Genetic and Evolutionary
  Computation}, pages 383--390. ACM.

\bibitem[Calma et~al., 2017]{CalmaKST17}
Calma, A., Kottke, D., Sick, B., and Tomforde, S. (2017).
\newblock Learning to learn: Dynamic runtime exploitation of various knowledge
  sources and machine learning paradigms.
\newblock In {\em Proc. 2nd {IEEE} Int. Workshops on Foundations and
  Applications of Self* Systems}, pages 109--116.

\bibitem[Chu et~al., 2011]{chu2011unbiased}
Chu, W., Zinkevich, M., Li, L., Thomas, A., and Tseng, B. (2011).
\newblock Unbiased online active learning in data streams.
\newblock In {\em Proc. of 17th ACM SIGKDD Int. Conf. on Knowledge discovery
  and data mining}, pages 195--203. ACM.

\bibitem[D'Angelo et~al., 2019]{DAngeloGGGNPT19}
D'Angelo, M., Gerasimou, S., Ghahremani, S., Grohmann, J., Nunes, I.,
  Pournaras, E., and Tomforde, S. (2019).
\newblock On learning in collective self-adaptive systems: state of practice
  and a 3d framework.
\newblock In {\em Proc of 14th SEAMS@ICSE}, pages 13--24.

\bibitem[D'Angelo et~al., 2020]{DAngeloGGGNTP20}
D'Angelo, M., Ghahremani, S., Gerasimou, S., Grohmann, J., Nunes, I., Tomforde,
  S., and Pournaras, E. (2020).
\newblock Learning to learn in collective adaptive systems: Mining design
  patterns for data-driven reasoning.
\newblock In {\em 2020 {IEEE} Int. Conf. on Autonomic Computing and
  Self-Organizing Systems, Companion}, pages 121--126.

\bibitem[Donmez et~al., 2007]{donmez2007dual}
Donmez, P., Carbonell, J.~G., and Bennett, P.~N. (2007).
\newblock Dual strategy active learning.
\newblock In {\em European Conference on Machine Learning}, pages 116--127.
  Springer.

\bibitem[Fredivianus et~al., 2012]{FredivianusKS12}
Fredivianus, N., Kara, K., and Schmeck, H. (2012).
\newblock Stay real!: {XCS} with rule combining for real values.
\newblock In {\em Genetic and Evolutionary Computation Conference}, pages
  1493--1494.

\bibitem[Fredivianus et~al., 2010]{FredivianusPS10}
Fredivianus, N., Prothmann, H., and Schmeck, H. (2010).
\newblock {XCS} revisited: {A} novel discovery component for the extended
  classifier system.
\newblock In {\em Simulated Evolution and Learning - 8th Int. Conf.}, pages
  289--298.

\bibitem[Glass, 2002]{Gla02}
Glass, R.~L. (2002).
\newblock {\em {Facts and Fallacies of Software Engineering}}.
\newblock Agile Software Development. Addison Wesley, Boston, US.

\bibitem[Gruhl et~al., 2021]{GruhlST21}
Gruhl, C., Sick, B., and Tomforde, S. (2021).
\newblock Novelty detection in continuously changing environments.
\newblock {\em Future Gener. Comput. Syst.}, 114:138--154.

\bibitem[Kangas et~al., 2014]{kangas2014efficient}
Kangas, J.~D., Naik, A.~W., and Murphy, R.~F. (2014).
\newblock Efficient discovery of responses of proteins to compounds using
  active learning.
\newblock {\em BMC bioinformatics}, 15(1):143.

\bibitem[Kephart and Chess, 2003]{KC03}
Kephart, J. and Chess, D. (2003).
\newblock The {V}ision of {A}utonomic {C}omputing.
\newblock {\em IEEE Computer}, 36(1):41--50.

\bibitem[Kounev et~al., 2017]{Kounev2017}
Kounev, S., Lewis, P., Bellman, K.~L., Bencomo, N., Camara, J., Diaconescu, A.,
  Esterle, L., Geihs, K., Giese, H., G{\"{o}}tz, S., Inverardi, P., Kephart,
  J.~O., and Zisman, A. (2017).
\newblock {The Notion of Self-aware Computing}.
\newblock In {\em Self-Aware Computing Systems}, pages 3--16. Springer.

\bibitem[Kovacs, 1999]{kovacs1999strength}
Kovacs, T. (1999).
\newblock Strength or accuracy? fitness calculation in learning classifier
  systems.
\newblock In {\em Int. Worksh. on Learning Classifier Sys.}, pages 143--160.
  Springer.

\bibitem[Lehman and Stanley, 2008]{lehman2008exploiting}
Lehman, J. and Stanley, K.~O. (2008).
\newblock Exploiting open-endedness to solve problems through the search for
  novelty.
\newblock In {\em ALIFE}, pages 329--336.

\bibitem[Lewis and Gale, 1994]{lewis1994sequential}
Lewis, D.~D. and Gale, W.~A. (1994).
\newblock A sequential algorithm for training text classifiers.
\newblock In {\em SIGIR’94}, pages 3--12. Springer.

\bibitem[Moore, 1965]{Moo65}
Moore, G.~E. (1965).
\newblock Cramming more components onto integrated circuits.
\newblock {\em Electronics Mag.}, 38(8):114 -- 117.

\bibitem[M\"uller-Schloer and Tomforde, 2017]{MST17}
M\"uller-Schloer, C. and Tomforde, S. (2017).
\newblock {\em Organic Computing -- Techncial Systems for Survival in the Real
  World}.
\newblock Autonomic Systems. Birkh\"auser Verlag.

\bibitem[Najar et~al., 2015a]{najar2015social}
Najar, A., Sigaud, O., and Chetouani, M. (2015a).
\newblock Social-task learning for hri.
\newblock In {\em Int. Conf. on Social Robotics}, pages 472--481. Springer.

\bibitem[Najar et~al., 2015b]{najar2015socially}
Najar, A., Sigaud, O., and Chetouani, M. (2015b).
\newblock Socially guided xcs: using teaching signals to boost learning.
\newblock In {\em Proc. of Companion to 2015 An. Conf. on Genetic and
  Evolutionary Comp.}, pages 1021--1028. ACM.

\bibitem[Nakata et~al., 2015]{nakata2015should}
Nakata, M., Lanzi, P.~L., Kovacs, T., Browne, W.~N., and Takadama, K. (2015).
\newblock How should learning classifier systems cover a state-action space?
\newblock In {\em Proc. of CEC15}, pages 3012--3019. IEEE.

\bibitem[Nissim et~al., 2014]{nissim2014novel}
Nissim, N., Moskovitch, R., Rokach, L., and Elovici, Y. (2014).
\newblock Novel active learning methods for enhanced pc malware detection in
  windows os.
\newblock {\em Expert Systems with Applications}, 41(13):5843--5857.

\bibitem[Orriols-Puig and Bernad{\'o}-Mansilla, 2009]{orriols2009evolutionary}
Orriols-Puig, A. and Bernad{\'o}-Mansilla, E. (2009).
\newblock Evolutionary rule-based systems for imbalanced data sets.
\newblock {\em Soft Computing}, 13(3):213.

\bibitem[Prothmann et~al., 2011]{PT+11}
Prothmann, H., Tomforde, S., Branke, J., H\"ahner, J., M\"uller-Schloer, C.,
  and Schmeck, H. (2011).
\newblock {Organic Traffic Control}.
\newblock In {\em Organic Computing -- A Paradigm Shift for Complex Systems},
  pages 431 -- 446. Birkh\"auser, Basel.

\bibitem[Reitmaier and Sick, 2013]{reitmaier2013let}
Reitmaier, T. and Sick, B. (2013).
\newblock Let us know your decision: Pool-based active training of a generative
  classifier with the selection strategy 4ds.
\newblock {\em Information Sciences}, 230:106--131.

\bibitem[Roy and McCallum, 2001]{RMC01}
Roy, N. and McCallum, A. (2001).
\newblock Toward optimal active learning through sampling estimation of error
  reduction.
\newblock In {\em Proc. of 18th Int. Conf. on Machine Learning}, pages
  441--448. Morgan Kaufman.

\bibitem[Rudolph et~al., 2019]{RudolphTH19}
Rudolph, S., Tomforde, S., and H{\"{a}}hner, J. (2019).
\newblock Mutual influence-aware runtime learning of self-adaptation behavior.
\newblock {\em {ACM} Trans. Auton. Adapt. Syst.}, 14(1):4:1--4:37.

\bibitem[Schweighofer and Doya, 2003]{schweighofer2003meta}
Schweighofer, N. and Doya, K. (2003).
\newblock Meta-learning in reinforcement learning.
\newblock {\em Neural Networks}, 16(1):5--9.

\bibitem[Settles, 2009]{settles2009active}
Settles, B. (2009).
\newblock Active learning literature survey.
\newblock Technical report, University of Wisconsin-Madison Department of
  Computer Sciences.

\bibitem[Settles, 2012]{settles2012active}
Settles, B. (2012).
\newblock Active learning.
\newblock {\em Synthesis Lect. on Art. Int. and Machine Learning}, 6(1):1--114.

\bibitem[Seung et~al., 1992]{seung1992query}
Seung, H.~S., Opper, M., and Sompolinsky, H. (1992).
\newblock Query by committee.
\newblock In {\em Proceedings of the fifth annual workshop on Computational
  learning theory}, pages 287--294. ACM.

\bibitem[Sigaud and Wilson, 2007]{sigaud2007learning}
Sigaud, O. and Wilson, S. (2007).
\newblock Learning classifier systems: a survey.
\newblock {\em Soft Comp.}, 11(11):1065--1078.

\bibitem[Stein et~al., 2017a]{stein2017toward}
Stein, A., Maier, R., and H{\"a}hner, J. (2017a).
\newblock Toward curious learning classifier systems: Combining xcs with active
  learning concepts.
\newblock In {\em GECCO17 Companion}, pages 1349--1356. ACM.

\bibitem[Stein et~al., 2016]{SteinRTH16}
Stein, A., Rauh, D., Tomforde, S., and H{\"{a}}hner, J. (2016).
\newblock Augmenting the algorithmic structure of {XCS} by means of
  interpolation.
\newblock In {\em Architecture of Computing Systems 2016}, pages 348--360.

\bibitem[Stein et~al., 2017b]{SRTH17}
Stein, A., Rauh, D., Tomforde, S., and H{\"a}hner, J. (2017b).
\newblock Interpolation in the extended classifier system: An architectural
  perspective.
\newblock {\em Journal of Systems Architecture}, 75:79--94.

\bibitem[Stein et~al., 2018]{SteinTDHM18}
Stein, A., Tomforde, S., Diaconescu, A., H{\"{a}}hner, J., and
  M{\"{u}}ller{-}Schloer, C. (2018).
\newblock A concept for proactive knowledge construction in self-learning
  autonomous systems.
\newblock In {\em Proc. of 3rd Int. Worksh. on Foundations and Applications of
  Self* Sys.}, pages 204--213.

\bibitem[Sutton and Barto, 1998]{SuttonB1998}
Sutton, R.~S. and Barto, A.~G. (1998).
\newblock {\em Introduction to Reinforcement Learning}.
\newblock MIT Press, 1st edition.

\bibitem[Tokic, 2010]{tokic2010adaptive}
Tokic, M. (2010).
\newblock Adaptive $\varepsilon$-greedy exploration in reinforcement learning
  based on value differences.
\newblock In {\em An. Conf. on Art. Int.}, pages 203--210. Springer.

\bibitem[Tomforde et~al., 2011a]{TB+11}
Tomforde, S., Brameshuber, A., H\"ahner, J., and M\"uller-Schloer, C. (2011a).
\newblock {Restricted On-line Learning in Real-world Systems}.
\newblock In {\em Proc. of CEC11}, pages 1628 -- 1635. IEEE.

\bibitem[Tomforde et~al., 2014]{THS14}
Tomforde, S., H{\"a}hner, J., and Sick, B. (2014).
\newblock {Interwoven Systems}.
\newblock {\em Informatik-Spektrum}, 37(5):483--487.
\newblock {Aktuelles Schlagwort}.

\bibitem[Tomforde et~al., 2010]{THH10-a}
Tomforde, S., Hurling, B., and H{\"a}hner, J. (2010).
\newblock {Dynamic control of mobile ad-hoc networks - Network protocol
  parameter adaptation using Organic Network Control}.
\newblock In {\em {Proc. of 7th Int. Conf. on Inf. in Control, Automation, and
  Robotics}}, pages 28--35. INSTICC.

\bibitem[Tomforde et~al., 2011b]{TP+11}
Tomforde, S., Prothmann, H., Branke, J., H\"ahner, J., Mnif, M.,
  M\"uller-Schloer, C., Richter, U., and Schmeck, H. (2011b).
\newblock {Observation and Control of Organic Systems}.
\newblock In {\em Organic Computing - A Paradigm Shift for Complex Systems},
  pages 325 -- 338. Birkh\"auser.

\bibitem[Tomforde et~al., 2017]{TomfordeSM17}
Tomforde, S., Sick, B., and M{\"{u}}ller{-}Schloer, C. (2017).
\newblock Organic computing in the spotlight.
\newblock {\em CoRR}, abs/1701.08125.

\bibitem[Tomforde et~al., 2009]{TS+09}
Tomforde, S., Steffen, M., H\"ahner, J., and M\"uller-Schloer, C. (2009).
\newblock Towards an {O}rganic {N}etwork {C}ontrol {S}ystem.
\newblock In {\em Proc. of the 6th ATC}, pages 2 -- 16. Springer.

\bibitem[Urbanowicz and Moore, 2015]{urbanowicz2015exstracs}
Urbanowicz, R. and Moore, J. (2015).
\newblock Exstracs 2.0: description and evaluation of a scalable learning
  classifier system.
\newblock {\em Ev. int.}, 8(2-3):89--116.

\bibitem[Urbanowicz et~al., 2012]{urbanowicz2012using}
Urbanowicz, R.~J., Granizo-Mackenzie, D., and Moore, J.~H. (2012).
\newblock Using expert knowledge to guide covering and mutation in a michigan
  style learning classifier system to detect epistasis and heterogeneity.
\newblock In {\em Int. Conf. on Parallel Problem Solving from Nature}, pages
  266--275. Springer.

\bibitem[Vermorel and Mohri, 2005]{vermorel2005multi}
Vermorel, J. and Mohri, M. (2005).
\newblock Multi-armed bandit algorithms and empirical evaluation.
\newblock In {\em ECML05}, pages 437--448. Springer.

\bibitem[Wilson, 1995]{Wilson1995}
Wilson, S.~W. (1995).
\newblock {Classifier Fitness Based on Accuracy}.
\newblock {\em Evolutionary Computation}, 3(2):149--175.

\bibitem[Wilson, 2000]{Wilson2000}
Wilson, S.~W. (2000).
\newblock Get real! xcs with continuous-valued inputs.
\newblock In {\em Learning Classifier Systems}, pages 209--219. Springer.

\bibitem[Wu and Miao, 2013]{wu2013curiosity}
Wu, Q. and Miao, C. (2013).
\newblock Curiosity: From psychology to computation.
\newblock {\em ACM Computing Surveys}, 46(2):18.

\end{thebibliography}

%
%

\end{document}